\documentclass[runningheads]{llncs}
\usepackage[T1]{fontenc}
\usepackage{graphicx}
\usepackage[utf8]{inputenc} 
\usepackage[T1]{fontenc}    
\usepackage{hyperref}       
\usepackage{url}            
\usepackage{booktabs}       
\usepackage{amsfonts}       
\usepackage{nicefrac}       
\usepackage{microtype}      
\usepackage{xcolor}         
 
\usepackage{algpseudocode} 
\usepackage{graphicx}

\usepackage{dsfont}
\usepackage{tikz}
\usepackage{pgfplots}
\usepackage{caption}
\usepackage{subcaption}
\usepackage{enumitem}
\usepackage{wrapfig}
\usepgfplotslibrary{fillbetween}
\usepackage[misc]{ifsym}

\newlength\figureheight
\newlength\figurewidth

\usepackage{bm}

\usepackage{amsthm,amsmath}
\usepackage{algorithm}

\begin{document}
\title{ENGAGE: Explanation Guided Data Augmentation for Graph Representation Learning}
\titlerunning{Explanation Guided Data Augmentation for Graph Learning}
%
\author{Yucheng Shi\inst{1} \and
Kaixiong Zhou\inst{2} \and
Ninghao Liu\inst{1}\Letter}
\authorrunning{Shi et al.}
%
\institute{University of Georgia, Athens, GA 30602, USA \\
\email{\{yucheng.shi, ninghao.liu\}@uga.edu}\and
Rice University, Houston, TX 77005, USA \\
\email{kaixiong.zhou@rice.edu}}
\maketitle   
\begin{abstract}
The recent contrastive learning methods, due to their effectiveness in representation learning, have been widely applied to modeling graph data.
Random perturbation is widely used to build contrastive views for graph data, which however, could accidentally break graph structures and lead to suboptimal performance. In addition, graph data is usually highly abstract, so it is hard to extract intuitive meanings and design more informed augmentation
schemes. Effective representations should preserve key characteristics in data and abandon superfluous information. 
In this paper, we propose ENGAGE (ExplaNation Guided data AuGmEntation), where explanation guides the contrastive augmentation process to preserve the key parts in graphs and explore removing superfluous information.
Specifically, we design an efficient unsupervised explanation method called smoothed activation map as the indicator of node importance in representation learning.
Then, we design two data augmentation schemes on graphs for perturbing structural and feature information, respectively. We also provide justification for the proposed method in the framework of information theories. Experiments of both graph-level and node-level tasks, on various model architectures and on different real-world graphs, are conducted to demonstrate the effectiveness and flexibility of ENGAGE. The code of ENGAGE can be found here \footnote[1]{{\color{blue} \url{https://github.com/sycny/ENGAGE}}.}.

\end{abstract}
\section{Introduction}
\label{introduction}
Graph representation learning has been shown to be powerful in many graph analysis tasks~\cite{hassani2020contrastive,zhu2020deep,you2021graph,you2020graph,shi2023chatgraph}. In particular, unsupervised graph representation learning methods~\cite{kipf2016variational,perozzi2014deepwalk,pan2018adversarially} have attracted substantial attention, as they are adaptive to various applications especially when labels are scarce. 
Among them, recently contrastive learning~\cite{tian2020makes,zhu2020deep} has achieved superior performance by learning representations from contrastive views to reduce superfluous information. 
However, most existing works~\cite{chen2020simple,chen2021exploring,zhu2020deep,you2020graph} simply apply random sampling for data augmentation, which could accidentally break the graph structures, thus reducing the effectiveness of representations. 
In other domains such as computer vision and natural language processing, more intelligent data augmentation methods have been proposed by leveraging the semantic information~\cite{peng2022crafting,tian2020makes} or domain knowledge~\cite{cheng2021fairfil}.
However, it is obscure to define and extract meaningful elements from abstract data types such as graphs. Also, the non-Euclidean property of graph increases the augmentation difficulty. 

Some preliminary efforts have been made to tackle the problem. For example, Fang et al.~\cite{fang2021molecular} propose using knowledge graphs to augment the original input graph and help construct contrastive pairs, which is applicable to computational chemistry. However, a more common and intriguing problem is how to design intelligent augmentation without external information sources. In addition, Zhu et al.~\cite{zhu2021graph} apply centrality as the node importance indicator and encourage perturbing nodes of lower centrality. However, node centrality is a static metric and is not necessarily relevant to downstream tasks in all scenarios. Some work~\cite{hassani2022learning,you2021graph} design end-to-end frameworks to automatically formulate data augmentation policies, but they could induce high computational costs. Also, their black-box nature would fail to explicitly locate task-relevant information in graph data. 

To solve the problem, this work proposes ENGAGE (ExplaNation Guided data AuGmEntation), where we use explanation to guide the generation of contrastive views for learning effective unsupervised representations on graph data. Specifically, we first propose a new explanation method called Smoothed Activation Map (SAM). Different from existing explanation methods that focus on understanding model predictions, SAM measures node importance based on the distribution of representations in the latent space.
Then, with the node importance scores, we design an explanation guided graph augmentation method by perturbing edges and node features, which is used to construct paired graph views for contrastive learning. By leveraging explanations, significant graph structures are less likely to be damaged by accidents. Meanwhile, our SAM method is efficient by using the quantization technique~\cite{johnson2019billion}, so that it can be applied into the training process. 
Finally, we conduct experiments on multiple datasets and tasks to demonstrate the effectiveness and flexibility of our proposed method.
The contributions can be summarized as below:
\begin{itemize}  
\item We propose the ENGAGE framework which leverages explanation to inform graph augmentation, and uses contrastive learning for training representations to preserve the key parts in graphs while removing uninformative artifacts.
\item We propose a new efficient explanation method called SAM for unsupervised representation learning on graph data.
\item We conduct comprehensive experiments on both node-level and graph-level classification tasks, with two representative contrastive learning models and different encoders, demonstrating the effectiveness and flexibility of ENGAGE.

\end{itemize}

\section{Related work}
\label{Related work}
\vspace{-4pt}

\subsection{Representation Learning for Graph Data}
Learning effective node or graph representations benefits various graph mining tasks. There are several major types of representation learning methods for graph data, including random-walk based methods~\cite{grover2016node2vec,perozzi2014deepwalk,qiu2018network,tang2015line}, graph neural networks (GNNs)~\cite{gao2019graph,hamilton2017inductive,velivckovic2018graph,velivckovic2018deep,wu2019simplifying,wu2023survey}, and graph contrastive learning~\cite{chen2020simple,zhu2020deep,xia2022simgrace,tan2023s2gae}. As the pioneering methodology of representation learning on graphs, random-walk based methods optimize node representations so that nodes within similar contexts tend to have similar representations. Then, graph neural networks become the new state-of-the-art architecture to process graph data. Finally, graph contrastive learning could be used to further refine the representation learning process with GNNs as the encoder.

\subsection{Graph Contrastive Learning}
Contrastive learning trains representations by maximizing the mutual information between different augmented views~\cite{chen2020simple,he2020momentum,chen2021exploring,tian2020contrastive,velivckovic2018deep}. 
For example, GRACE~\cite{zhu2020deep} constructs the negative node examples by dropping edges and masking features, and employs a contrastive model similar to SimCLR~\cite{chen2020simple}. GraphCL~\cite{you2020graph} perturbs graphs with four kinds of data augmentations on the graph level, and employs NT-Xent loss~\cite{chen2020simple} for training. SimGRACE~\cite{xia2022simgrace} gets rid of input data augmentation, and chooses to perturb model parameters.
Data augmentation is vital for contrastive learning to achieve good performance on downstream tasks~\cite{peng2022crafting,tian2020makes}. However, existing methods mainly rely on random perturbation, which could hurt graph information in the augmented views. 
To tackle the issue, Zhu et al.~\cite{zhu2021graph} propose to use node centrality as an importance indicator and perturb nodes with lower node centrality. 
Xu et al.~\cite{xu2021infogcl} propose an information-aware representation learning model~\cite{tishby2000information} to keep task-relevant information both in local and global views. Li et al.~\cite{Li2022LetIR} propose a graph rationale guided data augmentation to better preserve semantic information.
You et al.~\cite{you2021graph} design an end-to-end framework to automatically optimize data augmentation. 
However, it remains a challenge how to design data augmentation that is adaptive to different graph/node samples while keeping relatively low computing costs.

\subsection{Explanation for Graph Neural Networks}
Existing approaches for explaining GNN models can be divided into several categories~\cite{yuan2020explainability}, such as substitute-based, relevance-based, perturbation-based, generation-based, and rationale-based methods. Substitute-based methods approximate the original model with simplified but explainable substitutes, which either leverage GNN mechanisms~\cite{baldassarre2019explainability,pope2019explainability,wiltschko2020evaluating} or bypass the model information~\cite{huang2020graphlime,vu2020pgm,zhang2020relex}. Relevance-based methods compute input contributions by redistributing activations between neurons from the output layer to the input layer~\cite{pope2019explainability,schnake2020xai,schwarzenberg2019layerwise}. Perturbation-based methods~\cite{lucic2021cf,luo2020parameterized,schlichtkrull2020interpreting,wang2021causal,ying2019gnnexplainer} estimate input scores with the assumption that removing important features will have a large impact on the prediction. Generation-based methods produce synthetic graphs that maximally activate the target neuron~\cite{yuan2020xgnn}. Rationale-based methods find a small subgraph which best guides the model prediction as explanation~\cite{wu2022discovering,Li2022LetIR,Liu2022GraphRW}. 
Existing methods mainly focus on supervised graph learning models, while we attempt to obtain explanation for unsupervised graph learning. Meanwhile, we propose to leverage explanation for improving models, which further requires the explanation algorithm to be efficient.

\section{Preliminaries}
\label{Preliminaries}
\subsection{Notations}
Let $G = \left \{ \mathcal{V},\mathcal{E}, \bm{A}, \bm{X} \right \}$ be a graph, where $\mathcal{V}$ and $\mathcal{E}$ denote the set of nodes and edges, respectively. $\bm{A} \in \left\{0,1\right\}^{|\mathcal{V}|\times |\mathcal{V}|}$ is the adjacency matrix. If node $v_{i}$ and node $v_{j}$ are connected, then $A_{ij}=1$, where we use $e_{i,j}$ to denote that edge. $\bm{X}\in \mathbb{R}^{|{\mathcal{V}}|\times D}$ is the feature matrix and $\bm{x}_{i}  = \bm{X}_{i,:}$ is the feature vector of node $v_{i}$. The GNN encoder is denoted as $f(\cdot)$, which transforms the nodes to representations $\bm{Z} = f(\bm{X},\bm{A}), \bm{Z} \in \mathbb{R}^{\;|\mathcal{V}|\times 
K}$, where $K$ is the latent dimension and $\bm{z}_i=\bm{Z}_{i,:}$ denotes the representation of node $v_i$. Meanwhile, we also consider graph-level tasks in this work. Let $\mathcal{G}=\{G_{1},G_{2}, ..., G_{N}\}$ denote a set of graphs. In this case, the representation of a graph $G_n$ is obtained as $\bm{z}_n = f(G_n)$, where $\bm{z}_n\in \mathbb{R}^{K}$. Our proposed method is applicable to both types of tasks.
Unless otherwise stated, the methodology part in this paper assumes using graph-level tasks for illustration.

\subsection{Contrastive Learning Frameworks}
In this work, we adopt contrastive learning for learning node/graph representations. We consider two types of contrastive learning frameworks. The first type includes GraphCL~\cite{you2020graph} and GRACE~\cite{zhu2020deep} that are motivated by SimCLR~\cite{chen2020simple} and learn to maximize
the consistency between positive views compared with negative views. The second type is based on Simsiam~\cite{chen2021exploring} and could get rid of negative samples by using the stop gradient to prevent model collapse. We modify SimCLR and Simsiam frameworks to adapt to graph representation learning, and apply ENGAGE on both frameworks.

\subsubsection{Learning Objectives.}
In contrastive learning, each instance is augmented into two \textit{positive views} whose representations are $\bm{z}^1$ and $\bm{z}^2$, respectively. In SimCLR~\cite{chen2020simple}, the loss for learning the representation of $i$-th instance is:
\begin{equation}
\label{infornce}
\ell_i=-\log \frac{\exp \left(\operatorname{sim}\left(\bm{z}^1_{i}, \bm{z}^2_{i}\right) / \tau\right)}{\sum_{j} \mathds{1}_{[j \neq i]} \exp \left(\operatorname{sim}\left(\bm{z}^+_{i}, \bm{z}_{j}\right) / \tau\right)},
\end{equation}
where $\bm{z}^+_{i}$ refers to $\bm{z}^1_i$ or $\bm{z}^2_i$, and $\bm{z}_j$ denotes the embeddings of other instances as negative views. Positive views are expected to preserve the core information of the original instance, while negative views are expected to be irrelevant to the original instance. $\operatorname{sim}\left(\bm{z}^1, \bm{z}^2\right) = {\bm{z}^1}^T \bm{z}^2/\left \| \bm{z}^1 \right \| \left \| \bm{z}^2 \right \|$.
The Simsiam~\cite{chen2021exploring} model gets rid of the negative views, and its loss function is defined as below:
\begin{equation}
\label{simsiam}
    \ell_{i}=-(\bm{z}^1_i/\left\|\bm{z}^1_i\right\|_{2}) \cdot (\bm{z}^2_i/\left\|\bm{z}^2_i\right\|_{2}).
\end{equation}

\subsubsection{Encoder and MLP Heads.}
We use graph neural networks as the encoder $f(\cdot)$ to learn graph/node representations $\bm{z}$. To comprehensively test our proposed framework, we apply several architectures, including graph convolutional networks (GCN)~\cite{kipf2016semi} and graph attention networks (GAT)~\cite{velivckovic2018graph} for node-level tasks. Also, we employ graph isomorphism networks (GIN)~\cite{xu2018powerful} for graph-level tasks. Then, the representations are fed into different MLP heads depending on the CL framework. In SimCLR, there is only one MLP head called $p_{o}(\cdot)$. In Simsiam, there are two MLP heads, namely $p_{o}(\cdot)$ and $p_{e}(\cdot)$.

\section{The ENGAGE Framework}
\label{EPGDA}
\subsection{Mitigating Superfluous Information in Representations}
We begin by introducing the guideline of minimizing superfluous information, which we will follow throughout this work, for learning effective and robust representations. Without loss of generality, we denote the original input and downstream task label as $X$ and $\bm{y}$, respectively.
\begin{definition}
(Sufficiency) A representation $\bm{z}$ is defined as sufficient for label $\bm{y}$ iff  $I(X; \bm{y}) = I(\bm{z}; \bm{y})$.
\end{definition}

An illustration of sufficient representation is shown in Figure~\ref{views}(a).
To achieve good downstream task performance, it is encouraged to learn $\bm{z}$ from $X$ without losing task-relevant information.
Meanwhile, according to the information bottleneck principle~\cite{tishby2000information}, by reducing superfluous information from $X$, the sufficient representation $\bm{z}$ becomes more robust and gives better performance. This can be better understood by splitting the mutual information between $X$ and $\bm{z}$ into two parts as below:
\begin{equation}
\label{Ixz}
    I(X ; \bm{z})=\underbrace{I(X ; \bm{z} \mid \bm{y})}_{\text {superfluous information }}+\underbrace{I(\bm{z}; \bm{y})}_{\text {task-relevant information }} .
\end{equation}
The first term is the information in $\bm{z}$ that is not useful for predicting $\bm{y}$, which should be minimized. The second term denotes the predictive information, which is not affected by the representation as long as $\bm{z}$ is sufficient for $\bm{y}$. 

Recent research shows contrastive learning (CL) can control the amount of information preserved in representations~\cite{tian2020makes,wang2022rethinking}.
However, without label information, it is difficult to specify which part of input information to be preserved in CL.
In this work, we utilize explanation to identify non-trivial structural information in graphs. 
Then, the information recommended by explanation is given higher priority to be preserved in representations, while other information is more likely to be discarded. 
This is done by using explanation, instead of random perturbation, to guide data augmentation in contrastive learning.

\begin{figure}[t]
  \centering
  \includegraphics[width=0.99\textwidth]{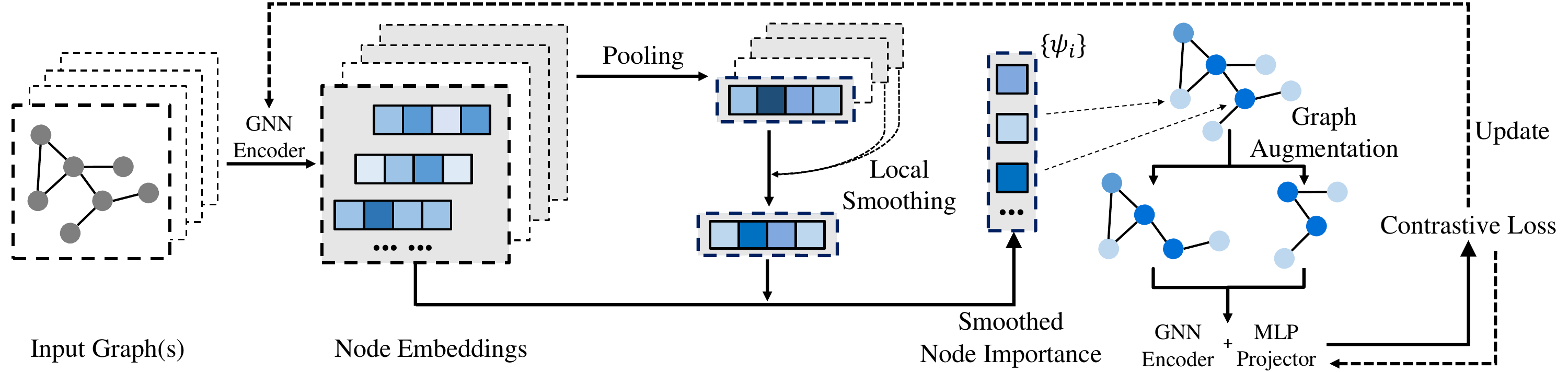}
  \caption{Graph augmentation guided by Smoothed Activation Map (SAM) for contrastive learning.}
  \label{explanation}
\end{figure}

\subsection{Efficient Explanations for Unsupervised Representations}
\subsubsection{Class Activation Map.} The goal of explanation is to identify the key components in input. We use GCN to illustrate the idea of explaining graph neural networks. A graph convolutional layer is defined as
$
    F^{l}=\sigma(\tilde{\bm{D}}^{-\frac{1}{2}} \tilde{\bm{A}}\tilde{\bm{D}}^{-\frac{1}{2}}F^{(l-1)} \bm{W}^{l}),
$
where $F^{l}$ is output of ${l}$-th layer, $F^{0}=\bm{X}$, $\tilde{\bm{A}} = \bm{A} +\bm{I}_{N}$ is the adjacency matrix with self-connections, and $\tilde{\bm{D}}$ is the degree matrix of $\tilde{\bm{A}}$. $\bm{W}^{l}$ denotes trainable weights, and $\sigma(\cdot)$ is the activation function. In supervised learning, the Class Activation Map (CAM)~\cite{pope2019explainability,zhou2016learning} can be used to compute the importance of node $v_i$ as:
$
    \psi_{i}^{c}=\operatorname{ReLU}(\textstyle\sum_{k} w^c_{k} F_{k, i}^{L}),
$
where $L$ is the final convolutional layer, and $F_{k,i}^{l}$ is the $k$-th latent dimension (i.e., channel) of node $v_{i}$ at the $l$-{th} layer. $w_{k}^{c}$ is the weight of channel $k$ at the output layer towards predicting class $c$.

However, there are two challenges that impede us from directly applying CAM to our problem. First, CAM is designed for specific model architectures with a GCN and a fully-connected output layer (consisting of the $w_{k}^{c}$ weights), where its applicability is limited. Second, CAM works for supervised models, while we focus on unsupervised learning.

\begin{wrapfigure}[12]{R}{0.44\textwidth}
\begin{minipage}{0.44\textwidth}
        \vspace{-8mm}
        \begin{subfigure}[t]{1.00\textwidth}
        \small
        \includegraphics[width=1.0\textwidth]{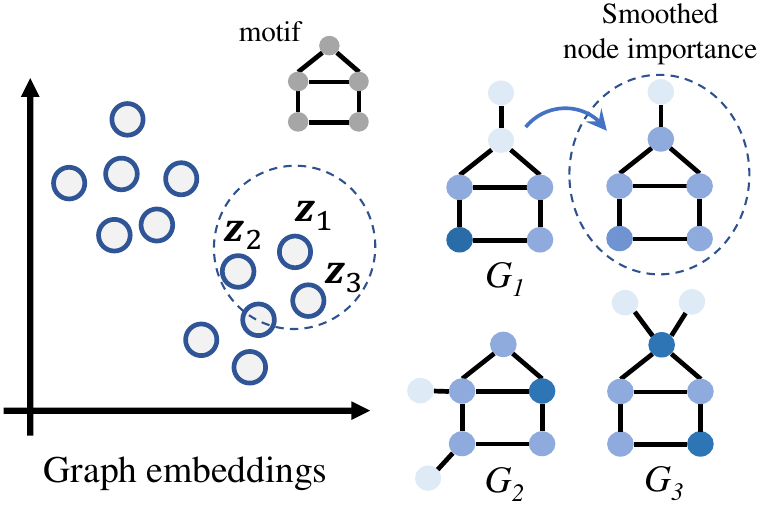}
        \vspace{-4mm}
        \end{subfigure} 
        \vspace{-4mm}
    \caption{Extracting reliable explanations via local smoothing.}
    \label{fig:interpretation}
\end{minipage}
\end{wrapfigure}

\subsubsection{Smoothed Activation Map.} To tackle the challenges, we propose Smoothed Activation Map (SAM) to explain representations without class labels.
The key idea is to leverage the distribution of graph/node representations to identify the locally important and reliable graph components, as shown in Figure~\ref{fig:interpretation}. 
Unlike CAM, our method works for different types of GNN models and does not require supervised signals. 
Without loss of generality, we write the feedforward process of GNNs as follows:
\begin{equation} 
    \begin{array}{c}
\mathbf{a}_{i}^{l}=\operatorname{AGGREGATION}^{l}(\{F_{i^{\prime}}^{(l-1)}: n^{\prime} \in \mathcal{N}_{i}\}), \,\,
F_{i}^{l}=\operatorname{COMBINE}^{l}(F_{i}^{(l-1)}, \mathbf{a}_{i}^{l}),
\end{array}
\end{equation}
where $F_{i}^{l}$ is the embedding of $v_i$ at $l$-th layer. $\mathcal{N}_{i}$ denotes the neighbors of $v_i$. Then, the SAM heat-map is calculated as: 
\begin{equation}\label{eq:sam}
    \psi_{i} = \operatorname{ReLU}({\textstyle\sum_{k}}\,\, \tilde{w}_{k}  F_{k, i}^{L}),
\end{equation}
where $\psi_{i}$ explains the importance score of $v_i$, $\tilde{w}_{k}$ is the smoothed importance of channel $k$, and $L$ is the final layer.
For graph-level tasks, the channel importance of graph $G_n$ is estimated based on its local context, so 
$
   \tilde{w}_{k}^{graph(n)} = \mathrm{Norm}(\sum_{n' \in \mathcal{\tilde{N}}_{n}} \mathrm{Pool}\{F_{i'}^{L}: i'\in G_{n'}\})[k],
$
where $\mathcal{\tilde{N}}_{n}$ denotes the set of neighbors graphs around $G_n$ in the embedding space. The $\mathrm{Pool()}$ operation produces the heat-map of $G_{n'}$ from its nodes embeddings, where we use Average Pooling in experiments. The $\mathrm{Norm()}$ operation means performing $L_2$ normalization for heat-maps. The set $\mathcal{\tilde{N}}_{n}$ of neighbors can be retrieved efficiently by using quantization techniques~\cite{johnson2019billion}.
For node-level tasks, the channel importance is estimated by averaging the heat-maps of nearby node embeddings, i.e.,
$
\tilde{w}_{k}^{node(i)} = \mathrm{Norm}(\sum_{i' \in \mathcal{\tilde{N}}_{i}} F_{i'}^{L})[k],
$
where $\mathcal{\tilde{N}}_{i}$ is the set of neighbors that are close to $v_i$ in the embedding space. The explanation process is shown in Figure~\ref{explanation}.

Both $\tilde{w}_{k}^{N}$ and $\tilde{w}_{k}^{G}$ are estimated in unsupervised learning, and they play a similar role as $w^c_{k}$ in CAM. By considering nearby nodes and graphs, explanation information of the target node/graph is smoothed. Another perspective to understand SAM is that it provides explanation not only for a single node or graph, but for a group of nodes or graphs in the local manifold region.

Finally, many explanation methods have been proposed for graph neural networks (e.g., perturbation-based, substitute-based, and generation-base methods), so we want to justify our selection of CAM-style explanation in this work. The main consideration is computational efficiency. The heat-maps in CAM-style methods are directly obtainable in the feedforward process of GNNs, requiring minimal additional computation. In contrast, both perturbation-based and substitute-based methods require sending a number of perturbed inputs to estimate the effect of different graph features, which leads to significantly greater time costs. Generation-based methods face the similar issue, as they require training a global explainer on a large number of graph samples.

\begin{algorithm}[t]
	\caption{SAM-based node importance estimation for graph-level tasks. } 
	\hspace*{\algorithmicindent} \textbf{Input:} {Encoder $f(\cdot)$, the target graph $G_n=\{\mathcal{V},\mathcal{E}, \bm{A}, \bm{X}\} \in \mathcal{G}$.} 
	\begin{algorithmic}[1]
	\State $\bm{Z} \gets f(\bm{X},\bm{A})$; $\bm{z} \gets \mathrm{Pool}(\bm{Z})$; \,\,\,\,{\color{lightgray}//\,$\bm{z}\in \mathbb{R}^{K}$, the graph-level embedding of $G_n$}
	\State For $G_n$, find its $m$ nearest-neighbor graphs $\tilde{\mathcal{N}}_{n}$ in the latent space;
	\State $\tilde{w}_{k}^{graph} \gets \mathrm{Norm}(\bm{z} + \sum_{n' \in \mathcal{\tilde{N}}_{n}}^{m} \bm{z}_{n'})[k] $; {\color{lightgray}//\,smoothed importance score}
		   \For{$v_i \in G_n $}
    			\State $F_{k, i}^{L} \gets \bm{Z}[i,k]$;
    			\State $\psi_{i} \gets \operatorname{ReLU}({\textstyle\sum_{k}}\,\, \tilde{w}_{k}^{graph(n)} F_{k, i}^{L})$; \,\,\,\,{\color{lightgray}//\,node importance score}
                \EndFor
	\end{algorithmic}  
	\hspace*{\algorithmicindent} \textbf{Output:} {Node importance scores $\{\psi_{i}\}$.}
\end{algorithm}

\subsection{Explanation-Guided Contrastive Views Generation}
We design two data augmentation operations utilizing explanation results, where graph components highlighted by explanation are considered as important~\cite{pope2019explainability,peng2022crafting}. The general idea is to keep important information intact, while maximally perturbing unimportant information.

\subsubsection{Edge Perturbation.} We define two masks $\bm{M}^{\mathrm{edge}, 1}, \bm{M}^{\mathrm{edge}, 2} \in \{0,1\}^{|\mathcal{V}|\times |\mathcal{V}|}$ to perturb edges and obtain different data views from the original graph. The adjacency matrix of the two new views are
\begin{equation}
\label{edgea2}
\bm{A}_{1} = \bm{A} \odot \bm{M}^{\mathrm{edge},1}, \,\,
\bm{A}_{2} = \bm{A} \odot \bm{M}^{\mathrm{edge},2},
\end{equation}
where $\odot$ denotes the Hadamard product. The masks are obtained based on the explanation results:
\begin{equation}
\label{edgea1}
M_{i,j}^{\mathrm{edge},1}=\begin{cases} 1,  &\text{if}\; \phi_{i,j}>\theta_{e} 
 \\ \mathrm{Bernoulli}(\phi_{i,j})  &\text{if}\; \phi_{i,j} \le \theta_{e} 
\end{cases}, \,
M_{i,j}^{\mathrm{edge},2}=\begin{cases} 1,  &\text{if}\; \phi_{i,j} >\theta_{e} 
 \\ 1-M_{i,j}^{\mathrm{edge},1}  &\text{if}\; \phi_{i,j} \le \theta_{e} 
 \end{cases},
\end{equation}
where $\theta_{e}$ is a threshold, and $\phi_{i,j} = \left(\psi_{i}+\psi_{j}\right)/2$ is the edge importance between $v_i$ and $v_j$. The edge importance averages the explanation scores of end nodes. The intuition of edge masking is that, with explanations, we want to maintain the important connections intact while perturbing the relatively unimportant part as much as possible. Based on the above definition, the two views are set to keep minimal mutual information while both contain task-relevant information.
Here $\theta_e$ is a threshold that controls the degree of distinction between the two graph views. Empirically, we set 
$
\theta_{e}=\mu_{\phi}+\lambda_{e}*\sigma_{\phi},
$
where $\mu_{\phi}$ and $\sigma_{\phi}$ are the mean value and standard deviation of all the edge interpretability from a graph or a batch of graphs. 
$\lambda_e$ is the hyperparameter used to control the size of important part, because the range of importance scores can be different for different datasets.

\subsubsection{Feature Perturbation.}
We also define two masks $\bm{M}^{\mathrm{feat}, 1}, \bm{M}^{\mathrm{feat}, 2} \in \{0,1\}^{|\mathcal{V}|\times D}$ to perturb node features. The feature matrix of the two views are
\begin{equation}
\label{nodea2}
    \bm{X}_{1} = \bm{X}\odot \bm{M}^{\mathrm{feat},1}, \,\,
    \bm{X}_{2} = \bm{X}\odot \bm{M}^{\mathrm{feat},2}.
\end{equation}
The masks are obtained based on explanation results:
\begin{equation}
\label{nodea1}
M_{i,d}^{\mathrm{feat},1}=\begin{cases} 1, \; &\text{if}\; \psi_{i} >\theta_{f} 
 \\ \mathrm{Bernoulli}(\psi_{i}) \; &\text{if}\; \psi_{i} \le \theta_{f},
\end{cases}, \,\,
M_{i,d}^{\mathrm{feat},2}=\begin{cases} 1, \; &\text{if}\; \psi_{i} >\theta_{f} 
 \\ 1-M_{i}^{\mathrm{feat},1} \; &\text{if}\; \psi_{i} \le \theta_{f}
\end{cases},
\end{equation}
where $\theta_{f}$ is a threshold, and $\psi_{i}$ is the explanation of node $v_i$.
Similar to edge perturbation, we set the threshold as $\theta_{f}=\mu_{\psi}+\lambda_{f}*\sigma_{\psi}$, where $\mu_{\psi}$ and $\sigma_{\psi}$ are the mean and standard deviation of node explanations from a graph or a batch of graphs. By this design, we can keep important information intact, while the features of unimportant nodes are more likely to be perturbed. Also, the two views are constructed in a way to reduce their mutual information $I(\bm{X}_1,\bm{X}_2)$.

\begin{figure}[t]
  \centering
  \includegraphics[width=0.9\textwidth]{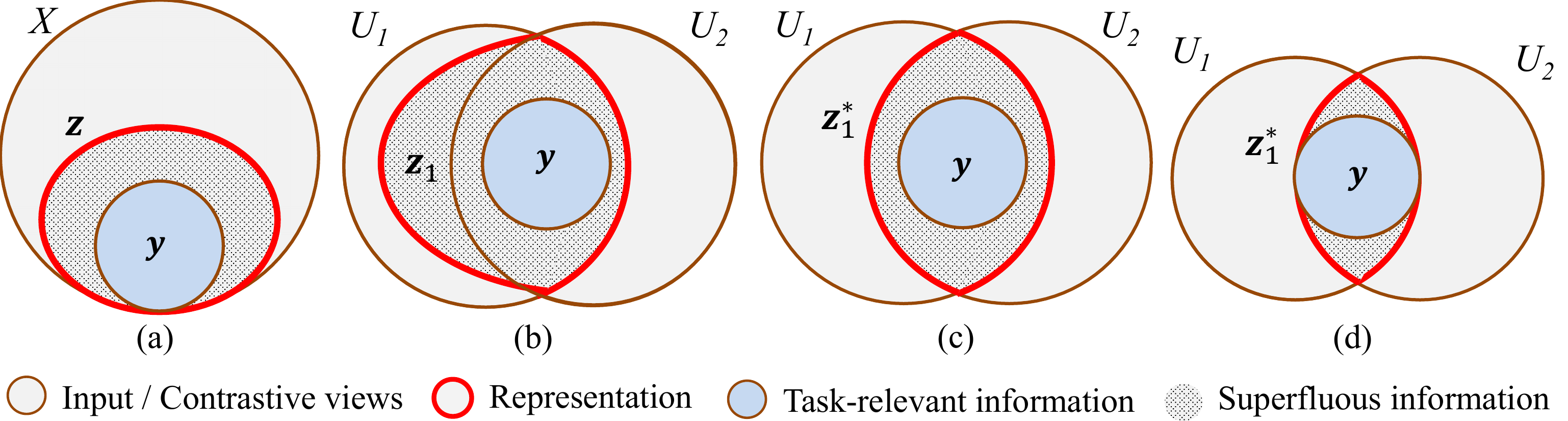}
  \caption{Illustration of theoretical justification (best viewed in color): (a) $\bm{z}$ contains all task-relevant information and is a sufficient representation of $X$; (b) $\bm{z}_1$ is a sufficient representation of $U_1$ containing task-relevant and non-shared information; (c)(d) $\bm{z}^*_1$ is the approximate minimal sufficient representation of $U_1$ that only includes shared information between $U_1$ and $U_2$.  }
  \label{views}
\end{figure}

\subsection{Theoretical Justification} 
In this part, we theoretically justify the design of our explanation guided method for building the contrastive views. The original input is denoted as $X$, the positive pair containing two data views obtained through augmentation is $\{U_{1},U_{2}\}$, and $\bm{y}$ is the downstream task label.

\begin{definition}
(Redundancy) The view $U_1$ is redundant with respect to view $U_2$ for $\bm{y}$ iff $I(U_1; \bm{y}| U_2)=0$. 
\end{definition}

The intuition behind is that a view $U_1$ is redundant for the task if the information in $\bm{y}$ is already observed in $U_2$. Let $\bm{z}_1$ and $\bm{z}_2$ denote the representation of $U_1$ and $U_2$, respectively. When $U_1$ and $U_2$ are mutually redundant, if $\bm{z}_1$ is sufficient for $U_2$ (refer to Definition~1 for "sufficiency"), then $\bm{z}_1$ retains task-relevant information about $\bm{y}$, as shown in Figure~\ref{views}(b). 
The proofs for the above statements are provided in Appendix A\footnote[2]{The appendix file is provided here: \url{https://github.com/sycny/ENGAGE}.}.

The statements above suggest that, in contrastive learning, the task-relevance of the representations $\bm{z}_1$ and $\bm{z}_2$ depends on the quality of the constructed contrastive views $U_1$ and $U_2$, where it is crucial to keep important task-relevant information intact when building $U_1$ and $U_2$. However, in unsupervised learning, the task label is not accessible. Thus, our assumption in this work is that, the graph information important for $\bm{z}$ in unsupervised representation distribution could also be important for downstream tasks, and should be preserved in $U_1$ and $U_2$ with higher priority. This motivates the design of explanation guided contrastive views construction from Equation~~\ref{edgea2}$\sim$~\ref{nodea1}, since graph components with high importance scores could be regarded as approximately preserving the key information of graphs. 
In addition, to mitigate noises in a single node/graph instance, our proposed SAM method gathers the explanatory information from the local context.

\begin{definition}
(Sufficient Contrastive Representation) A representation $\bm{z}_1$ of $U_1$ is sufficient for $U_2$ iff $I(\bm{z}_1; U_2) = I(U_1; U_2)$.
\end{definition}
\begin{definition}
(Minimal Sufficient Contrastive Representation) The sufficient contrastive representation $\bm{z}_1$ of $U_1$ is defined as minimal iff  $I(\bm{z}_1; U_1) \le I(\bm{z}'_1; U_1)$, $\forall \bm{z}'_1$ that is a sufficient contrastive representation.
\end{definition}

Let $\bm{z}^*_1$ denote the minimal sufficient contrastive representation learned from $U_1$. We can have $I(\bm{z}^*_1; U_2) = I(U_1; U_2)$, which implies that the shared information $I(U_1; U_2)$ is retained in the contrastive representation while the non-shared information is ignored~\cite{federici2020learning}. To obtain such representations, the encoder in contrastive learning is trained so that $\bm{z}_1\approx \bm{z}^*_1$ and $\bm{z}_2\approx \bm{z}^*_2$ (it also implies $I(\bm{z}_1, \bm{z}_2) \approx I(U_1, U_2)$)~\cite{wang2022rethinking}. After encoding, the minimal sufficient representation $\bm{z}_1$ (or $\bm{z}_2$) is obtained from $U_1$ (or $U_2$), and is used for prediction in downstream tasks. An illustration of $\bm{z}^*_1$ is presented in Figure~\ref{views}(c). However, the representation may still contain a significant amount of superfluous information, which could degrade the downstream task performance. 

Our design in Equation~~\ref{edgea2}$\sim$~\ref{nodea1} aims to address the above problem.
In our ENGAGE method, when thresholds $\theta_e$ and $\theta_f$ are set higher in Equation~\ref{edgea2} and~\ref{nodea2}, more information in $U_1$ and $U_2$ is deleted, making $I(U_1, \bm{z}_1|\bm{y})$ further reduced after training for $\bm{z}_1$. Thus, assuming $I(\bm{z}; \bm{y})$ is untouched (i.e., $U_1$ and $U_2$ remain mutually redundant) by setting appropriate thresholds, our proposed method reduces $I(X; \bm{z})$ by suppressing the superfluous information contained in $\bm{z}$, as depicted in Figure~\ref{views}(d). On the other hand, when $\theta_e$ and $\theta_f$ values are set smaller, we keep more information in $\bm{z}$ after contrastive learning.

\section{Experiments}
\label{Evaluation}
We answer the following research questions in experiments. 
\textbf{RQ1}: How effective is the proposed ENGAGE method in building contrastive views and learning graph representations?
\textbf{RQ2}: How does the proposed method influence representation learning as indicated by explanations? 
\textbf{RQ3}: What is the effect of hyperparameters (e.g., $\lambda_e$ and $\lambda_f$) on the proposed method?

\subsection{Experimental Setup}
We apply ENGAGE to contrastive learning models for both unsupervised graph-level and node-level classification tasks. For graph classification, we follow the evaluation pipeline in~\cite{sun2019infograph} by training an SVM as the classifier. The whole dataset is used to learn graph-level representations, which are evaluated under the SVM classifier with cross-validation. Eight benchmark datasets are all selected from the TUDataset. We modify SimCLR and Simsiam to adapt to graph representation learning and select GIN~\cite{xu2018powerful} as the encoder network. Besides using contrastive learning models with random perturbation as baselines, we compare our proposed model with graph kernel methods like WL~\cite{shervashidze2011weisfeiler} and DGK~\cite{yanardag2015deep}. We also choose state-of-the-art graph self-supervised learning methods including GraphCL~\cite{you2020graph}, JOAO~\cite{you2021graph}, JOAO(v2)~\cite{you2021graph}, SimGRACE~\cite{xia2022simgrace} and MVGRL~\cite{hassani2020contrastive}. 

For node classification, we follow~\cite{zhu2020deep} where the whole dataset is used to learn node representations, which are evaluated under a logistic regression classifier with cross-validation. Popular datasets like Cora and CiteSeer are selected, and other real-world datasets, including Wiki-CS, Amazon-Computers, and Amazon-Photo are also used. Similar to graph-level tasks, we modify SimCLR and Simsiam frameworks to adapt to node representation learning. Both GCN~\cite{kipf2016semi} and GAT~\cite{velivckovic2018graph} models are used as encoders.
We also compare with other self-supervised learning models including GCA~\cite{zhu2021graph}, MVGRL~\cite{hassani2020contrastive}, DGI~\cite{velivckovic2018deep}. We evaluate model performance using the \textit{accuracy} metric. For each model, we report mean accuracy and standard deviation of five runs. The `---' in Table~\ref{graph_classificaiotn} and Table~\ref{node_classificaiotn} means that these results are not available in currently published papers. The implementation details of our experiments can be found in Appendix G. 

\begin{table}[t]
  \renewcommand\arraystretch{1.3}
  \caption{Graph classification performance comparison.}
  \label{graph_classificaiotn}
  \centering
\resizebox{0.99\columnwidth}{!}{
\begin{tabular}{l|cccc|cccc|c}
\hline
Method          & NCI1                  & PROTEINS              & DD         &PTC-MR        & COLLAB      & RDT-B       & RDT-M5K     & IMDB-B & A.R.\\\hline\hline
WL              & 80.01 ± 0.50          &72.92 ± 0.56           &74.02 ± 2.28     &58.00 ± 0.50                &69.30 ± 3.44                   &68.82 ± 0.41          &46.06 ± 0.21       &\textbf{72.30 ± 3.44} &9.62\\
DGK          &\textbf{80.31 ± 0.46}           &73.30 ± 0.82           &74.85 ± 0.74    &60.10 ± 2.60                 &64.66 ± 0.50                   &78.04 ± 0.39          &41.27 ± 0.18          &66.96 ± 0.56 &9.88\\ \hline
node2vec              & 54.89 ± 1.61          &57.49 ± 3.57           &---   &58.60 ± 8.00                   &56.10 ± 0.20                  &---          &---       &--- &6.25\\
sub2vec             &52.84 ± 1.47           &53.03 ± 5.55           &54.33 ± 2.44      &60.00 ± 6.40               &55.26 ± 1.54                   &71.48 ± 0.41           &36.68 ± 0.42          &55.26 ± 1.54 &13.12\\
graph2vec              & 73.22 ± 1.81          &73.30 ± 2.05           &70.32 ± 2.32   &60.20 ± 6.90                  &71.10 ± 0.54                   &75.78 ± 1.03          &47.86 ± 0.26        &71.10 ± 0.54 &9.62\\
MVGRL             &77.00 ± 0.80 &---   &--- &\textbf{62.50 ± 1.70}   &\textbf{76.00 ± 1.20}   &84.50 ± 0.60  &--- &\textbf{74.20 ± 0.70} &\textbf{3.12}\\
InfoGraph             &76.20 ± 1.06           &74.44 ± 0.31           &72.85 ± 1.78  &\textbf{61.70 ± 1.40}                   &70.65 ± 1.13                   &82.50 ± 1.42           &53.46 ± 1.03          &\textbf{73.03 ± 0.87} &7.62\\
GraphCL         &77.87 ± 0.41           &74.39 ± 0.45           &\textbf{78.62 ± 0.40} &61.30 ± 2.10  &71.36 ± 1.15          &\textbf{89.53 ± 0.84} &55.99 ± 0.28 &71.14 ± 0.44 &5.38\\
JOAO            &78.07 ± 0.47           &74.55 ± 0.41           &77.32 ± 0.54    &---       &69.50 ± 0.36          &85.29 ± 1.35          &55.74 ± 0.63          &70.21 ± 3.08 &6.75\\
JOAOv2          &78.36 ± 0.53           &74.07 ± 1.10           &77.40 ± 1.15    &---        &69.33 ± 0.34          &86.42 ± 1.45          &\textbf{56.03 ± 0.27}          &70.83 ± 0.25 &6.12\\
SimGRACE        &79.12 ± 0.44  &75.35 ± 0.09  &77.44 ± 1.11   &---        &71.72 ± 0.82 &\textbf{89.51 ± 0.89} &55.91 ± 0.34 &71.30 ± 0.77 &4.00\\
\hline \hline
RD-SimCLR        & 79.02 ± 0.52          &74.61 ± 0.56           &77.40 ± 0.81 &58.55 ± 2.01    &69.32 ± 1.54     &85.67 ± 5.40          &55.52 ± 0.74       &69.08 ± 2.47  &8.00\\
EG-SimCLR        &\textbf{82.97 ± 0.20}           &\textbf{75.44 ± 0.65}     &\textbf{78.86 ± 0.51}     &61.51 ± 2.41                     &\textbf{76.60 ± 1.26}                   &\textbf{90.70 ± 0.46}          &\textbf{56.22 ± 0.56}       &71.78 ± 0.34 &\textbf{2.12} \\\hline
RD-Simsiam        &79.62 ± 0.59          &\textbf{75.42 ± 0.50}           &77.20 ± 1.33    &58.55 ± 	1.85              &66.50 ± 2.31 &86.12 ± 1.80          &54.30 ± 0.64       &69.86 ± 0.86  &7.88 \\
EG-Simsiam        & \textbf{81.49 ± 0.19}        &\textbf{76.06 ± 0.51}         &\textbf{78.35 ± 0.83}          &\textbf{62.67 ± 1.50}           &\textbf{74.73 ± 0.79}                   &89.17 ± 0.43          &\textbf{56.37 ± 0.17}      &71.93 ± 0.40 &\textbf{2.38} \\ \hline
\end{tabular}}
\end{table}

\subsection{Experiment Results and Comparisons}
\label{experiments}
Table~\ref{graph_classificaiotn} lists the result for graph classification, where the top three results are highlighted in bold. The proposed ENGAGE models are abbreviated as EG-SimCLR and EG-Simsiam, corresponding to their vanilla versions RD-SimCLR and RD-Simsiam, respectively. The results show that the average ranks (A.R.) of EG-SimCLR and EG-Simsiam are highest as 2.12 and 2.38, respectively, compared with other state-of-the-art methods. Among them, NCI1 has the biggest improvement of 2.66\% over the most competitive baselines. Most importantly, ENGAGE methods consistently outperform random perturbation based models (e.g., RD-SimCLR and RD-Simsiam) on the eight datasets, with an average improvement of 2.90\%.  In addition, ENGAGE reduces performance variance compared with random perturbation, indicating that representation learning becomes more stable. These observations validate the effectiveness of using explanation toward a more adaptive and intelligent data augmentation scheme.

The result for the node classification is presented in Table~\ref{node_classificaiotn} with the top three results highlighted. Our ENGAGE models are named as EG-SimCLR and EG-Simsiam, while their vanilla counterparts are GRACE~\cite{zhu2020deep} and RD-Simsiam, respectively. Many observations are similar to graph classification, such as improved performance and representation learning stability.
In addition, we observe that ENGAGE works better for the GCN architecture than GAT. The reason could be that the attention mechanism in GAT already equips it with some abilities to select graph components during training. The proposed ENGAGE is compatible with various GNN backbones and contrastive learning models.

\begin{table}[t]
  \renewcommand\arraystretch{1.3}
  \caption{Node classification performance comparison.}
  \label{node_classificaiotn}
  \centering
\resizebox{0.99\columnwidth}{!}{
\begin{tabular}{l|ccccc|c}
\hline
Method             & Cora          & Citeseer         &Wiki-CS  &Amazon-Computers  &Amazon-Photo &A.R\\ \hline
\hline
GCN            &81.50 ± 0.00       &70.30 ± 0.00      &77.19 ± 0.12     &86.51 ± 0.54  &92.42 ± 0.22 
 &11.6\\
GAT            &83.00 ± 0.70 &\textbf{72.50 ± 0.70}   &77.65 ± 0.11   &86.93 ± 0.29  &92.56 ± 0.35 &7.2\\
\hline
DeepWalk       &70.70 ± 0.60   &51.40 ± 0.50   &77.21 ± 0.03 &86.28 ± 0.07  &90.05 ± 0.08 &13.6 \\
GAE            &71.50 ± 0.40  &65.80 ± 0.40  &70.15 ± 0.01 &85.27 ± 0.19  &91.62 ± 0.13 & 13.6\\
DGI             &82.30 ± 0.60 &71.80 ± 0.70  &75.35 ± 0.14   &83.95 ± 0.47   &91.61 ± 0.22 &11.4 \\
MVGRL           &\textbf{86.80 ± 0.50} &\textbf{73.30 ± 0.50} &77.52 ± 0.08   &87.52 ± 0.11 &91.74 ± 0.07 &6.8\\ 
GCA             &---        &--- &78.35 ± 0.05  &87.85 ± 0.31  &92.53 ± 0.16 &8.3 \\
\hline\hline
GRACE             & 82.01 ± 0.56          &71.51 ± 0.33         &\textbf{79.12 ± 0.15} &88.36 ± 0.18 &92.52 ± 0.26 &6.8\\
EG-SimCLR$(GCN)$     &\textbf{84.07 ± 0.18}  &\textbf{72.40 ± 0.46} &\textbf{79.21 ± 0.12}  &\textbf{88.53 ± 0.13}  &92.65 ± 0.19  &\textbf{3.0}\\ 
\hline
    GRACE$(GAT)$     &83.57 ± 0.55       &71.70 ± 0.55        &78.48 ± 0.18 &88.09 ± 0.15 &\textbf{92.74 ± 0.20} &5.4\\
EG-SimCLR$(GAT)$     &\textbf{83.78 ± 0.53}  &72.28 ± 0.40 &78.76 ± 0.07   &\textbf{88.56 ± 0.17}  &\textbf{92.97 ± 0.07}  &\textbf{3.2}\\
\hline
RD-Simsiam$(GCN)$      &82.65 ± 0.62 &70.72 ± 0.64 &78.79 ± 0.10  &87.40 ± 0.31 &92.53 ± 	0.17 &8.2\\
EG-Simsiam$(GCN)$    &83.46 ± 0.56  &71.49 ± 0.31  &\textbf{79.27 ± 0.34}   &\textbf{88.66 ± 0.19}  &\textbf{92.77 ± 0.14} &\textbf{3.4}\\
\hline
RD-Simsiam$(GAT)$       &80.31 ± 0.34 &70.18 ± 0.54 &78.61 ± 0.27  &87.96 ± 0.27 &92.63 ± 	0.12 &8.8\\
EG-Simsiam$(GAT)$     &82.31 ± 0.43  &70.73 ± 0.17 &78.93 ± 0.20   &88.28 ± 0.18  &92.73 ± 0.28 &6.0\\ \hline
\end{tabular}}
\end{table}

\subsection{Ablation Study}
We conduct ablation studies to further verify the effect of using SAM explanations for guiding graph representation learning, and the influence of hyperparameters.

\subsubsection{Does ENGAGE remove redundant information?}
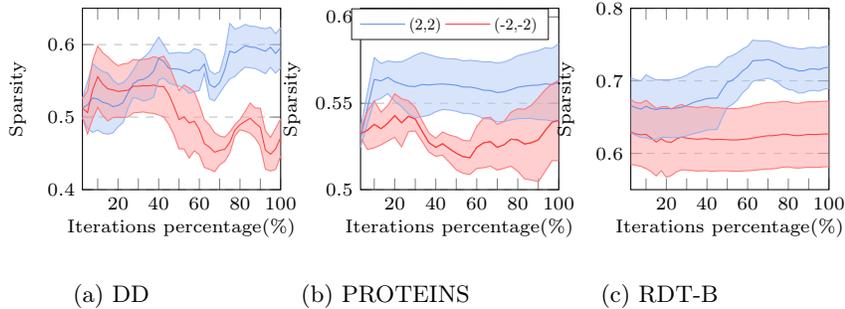
\begin{figure}[t]
        \begin{subfigure}{0.24\textwidth}
        \setlength \figureheight{0.95in}
        \setlength \figurewidth{1.4in}
        \centering  \scriptsize
        \definecolor{mycolor1}{rgb}{1,0.65,0}%
\definecolor{mycolor2}{rgb}{1,0.94,0}%
\definecolor{mycolor3}{rgb}{1.0, 0.25, 0.25}%
\definecolor{mycolor4}{rgb}{0.39, 0.58, 0.93}
\definecolor{mycolor5}{rgb}{0., 0.18, 0.39}

\begin{tikzpicture}
    \begin{axis}[
    width=0.951\figurewidth,
    height=\figureheight,
    at={(0\figurewidth,0\figureheight)},
    scale only axis,
    scaled x ticks=true,
    xticklabels={0,,20,,40,,60,,80,,100},
    xtick={0,4,8,12,16,20,24,28,32,36,40},
    x label style={at={(axis description cs:0.5,0.1)},anchor=north},
    xlabel style={font=\color{white!15!black},font=\fontsize{7.5}{7.5}\selectfont},
    xlabel={Iterations percentage(\%)},
    y label style={at={(axis description cs:0.22,0.5)},anchor=south},
    ylabel style={font=\color{white!15!black},,font=\fontsize{7.5}{7.5}\selectfont},
    ylabel={Sparsity},
      xmin=1, xmax=40,
      ymin=0.4, ymax=0.65,
      ymajorgrids=true,
      grid style=dashed,
      legend columns = 2,
      legend style={font=\fontsize{10}{10}\selectfont, legend cell align=left, align=left, draw=white!15!black, nodes={scale=0.7}, at={(0.85, 1.2)}},
      width=0.9\textwidth,
    ]

    \addplot[color=mycolor4] coordinates {(1,0.5126)(2,0.5168)(3,0.5264)(4,0.5248)(5,0.5203)(6,0.5186)(7,0.5145)(8,0.5160)(9,0.5200)(10,0.5337)(11,0.5461)(12,0.5509)(13,0.5521)(14,0.5561)(15,0.5633)(16,0.5809)(17,0.5767)(18,0.5673)(19,0.5665)(20,0.5660)(21,0.5639)(22,0.5606)(23,0.5647)(24,0.5644)(25,0.5732)(26,0.5448)(27,0.5410)(28,0.5477)(29,0.5654)(30,0.5910)(31,0.5886)(32,0.5934)(33,0.5983)(34,0.5967)(35,0.5960)(36,0.5933)(37,0.5909)(38,0.5963)(39,0.5878)(40,0.5956)};
    
    \addplot[color=mycolor3] coordinates {(1,0.5124)(2,0.5063)(3,0.5376)(4,0.5556)(5,0.5454)(6,0.5389)(7,0.5377)(8,0.5354)(9,0.5363)(10,0.5372)(11,0.5426)(12,0.5426)(13,0.5433)(14,0.5431)(15,0.5442)(16,0.5423)(17,0.5421)(18,0.5313)(19,0.5149)(20,0.4965)(21,0.5002)(22,0.4919)(23,0.4944)(24,0.4828)(25,0.4634)(26,0.4586)(27,0.4516)(28,0.4541)(29,0.4557)(30,0.4652)(31,0.4820)(32,0.4896)(33,0.4942)(34,0.4985)(35,0.4918)(36,0.4859)(37,0.4547)(38,0.4488)(39,0.4561)(40,0.4720)};
    
    \addplot[name path=us_top,color=mycolor4!70] coordinates {(1,0.5257)(2,0.5489)(3,0.5738)(4,0.5674)(5,0.5630)(6,0.5607)(7,0.5508)(8,0.5454)(9,0.5475)(10,0.5545)(11,0.5648)(12,0.5767)(13,0.5822)(14,0.5881)(15,0.5991)(16,0.6102)(17,0.5996)(18,0.5877)(19,0.5884)(20,0.5892)(21,0.5884)(22,0.5884)(23,0.5945)(24,0.5922)(25,0.5863)(26,0.5642)(27,0.5595)(28,0.5625)(29,0.5873)(30,0.6291)(31,0.6212)(32,0.6238)(33,0.6277)(34,0.6254)(35,0.6250)(36,0.6230)(37,0.6219)(38,0.6240)(39,0.6151)(40,0.6232)};
    
    \addplot[name path=us_down,color=mycolor4!70] coordinates {(1,0.4994)(2,0.4847)(3,0.4789)(4,0.4821)(5,0.4777)(6,0.4765)(7,0.4782)(8,0.4866)(9,0.4925)(10,0.5129)(11,0.5273)(12,0.5252)(13,0.5221)(14,0.5240)(15,0.5274)(16,0.5515)(17,0.5538)(18,0.5469)(19,0.5445)(20,0.5428)(21,0.5393)(22,0.5328)(23,0.5349)(24,0.5366)(25,0.5602)(26,0.5254)(27,0.5224)(28,0.5329)(29,0.5435)(30,0.5528)(31,0.5561)(32,0.5630)(33,0.5689)(34,0.5679)(35,0.5670)(36,0.5636)(37,0.5599)(38,0.5687)(39,0.5605)(40,0.5680)};
    
    \addplot[mycolor4!50,fill opacity=0.5] fill between[of=us_top and us_down];

    \addplot[name path=us_top,color=mycolor3!70] coordinates {(1,0.5256)(2,0.5358)(3,0.5777)(4,0.5979)(5,0.5889)(6,0.5830)(7,0.5760)(8,0.5709)(9,0.5732)(10,0.5743)(11,0.5794)(12,0.5786)(13,0.5796)(14,0.5804)(15,0.5833)(16,0.5816)(17,0.5840)(18,0.5727)(19,0.5542)(20,0.5367)(21,0.5442)(22,0.5402)(23,0.5351)(24,0.5174)(25,0.4962)(26,0.4892)(27,0.4784)(28,0.4740)(29,0.4723)(30,0.4795)(31,0.4899)(32,0.4968)(33,0.5078)(34,0.5194)(35,0.5125)(36,0.5123)(37,0.4811)(38,0.4719)(39,0.4845)(40,0.4977)};
    \addplot[name path=us_down,color=mycolor3!70] coordinates {(1,0.4991)(2,0.4768)(3,0.4975)(4,0.5133)(5,0.5019)(6,0.4948)(7,0.4993)(8,0.4999)(9,0.4994)(10,0.5001)(11,0.5058)(12,0.5065)(13,0.5071)(14,0.5058)(15,0.5051)(16,0.5030)(17,0.5002)(18,0.4898)(19,0.4755)(20,0.4562)(21,0.4561)(22,0.4436)(23,0.4536)(24,0.4482)(25,0.4306)(26,0.4281)(27,0.4247)(28,0.4342)(29,0.4390)(30,0.4508)(31,0.4741)(32,0.4825)(33,0.4806)(34,0.4777)(35,0.4710)(36,0.4596)(37,0.4283)(38,0.4257)(39,0.4278)(40,0.4463)};
    \addplot[mycolor3!50,fill opacity=0.5] fill between[of=us_top and us_down];
    
    \end{axis}
\end{tikzpicture} 
            \caption{{\small DD}}  
        \end{subfigure}
        \hspace{0.5cm}
        \begin{subfigure}{0.24\textwidth}
        \setlength 
        \figureheight{0.95in}
        \setlength 
        \figurewidth{1.4in}
        \centering  \scriptsize
        \definecolor{mycolor1}{rgb}{1,0.65,0}%
\definecolor{mycolor2}{rgb}{1,0.94,0}%
\definecolor{mycolor3}{rgb}{1.0, 0.25, 0.25}%
\definecolor{mycolor4}{rgb}{0.39, 0.58, 0.93}
\definecolor{mycolor5}{rgb}{0., 0.18, 0.39}

\begin{tikzpicture}
    \begin{axis}[
    width=0.951\figurewidth,
    height=\figureheight,
    at={(0\figurewidth,0\figureheight)},
    scale only axis,
    scaled x ticks=true,
    xticklabels={0,,20,,40,,60,,80,,100},
    xtick={0,3,6,9,12,15,18,21,24,27,30},
    x label style={at={(axis description cs:0.5,0.1)},anchor=north},
    xlabel style={font=\color{white!15!black},,font=\fontsize{7.5}{7.5}\selectfont},
    xlabel={Iterations percentage(\%)},
    y label style={at={(axis description cs:0.2,0.5)},anchor=south},
    ylabel style={font=\color{white!15!black},,font=\fontsize{7.5}{7.5}\selectfont},
    ylabel={Sparsity},
      xmin=1, xmax=30,
      ymin=0.50, ymax=0.605,
      ymajorgrids=true,
      grid style=dashed,
      legend columns = 2,
      legend style={font=\fontsize{10}{10}\selectfont, legend cell align=left, align=left, draw=white!15!black, nodes={scale=0.6}, at={(0.95, 1)}},
      width=0.9\textwidth,
    ]

    \addplot[color=mycolor4] coordinates {(1,0.5299)(2,0.5471)(3,0.5638)(4,0.5630)(5,0.5650)(6,0.5625)(7,0.5611)(8,0.5599)(9,0.5597)(10,0.5609)(11,0.5606)(12,0.5610)(13,0.5612)(14,0.5609)(15,0.5605)(16,0.5594)(17,0.5594)(18,0.5587)(19,0.5576)(20,0.5574)(21,0.5562)(22,0.5567)(23,0.5579)(24,0.5593)(25,0.5599)(26,0.5604)(27,0.5605)(28,0.5613)(29,0.5608)(30,0.5624)};
    
    \addplot[color=mycolor3] coordinates {(1,0.5324)(2,0.5344)(3,0.5378)(4,0.5353)(5,0.5385)(6,0.5429)(7,0.5387)(8,0.5425)(9,0.5409)(10,0.5339)(11,0.5272)(12,0.5245)(13,0.5266)(14,0.5237)(15,0.5211)(16,0.5191)(17,0.5185)(18,0.5245)(19,0.5271)(20,0.5278)(21,0.5245)(22,0.5260)(23,0.5292)(24,0.5288)(25,0.5264)(26,0.5272)(27,0.5288)(28,0.5341)(29,0.5390)(30,0.5401)};

    \addplot[name path=us_top,color=mycolor4!70] coordinates {(1,0.5355)(2,0.5561)(3,0.5764)(4,0.5731)(5,0.5745)(6,0.5739)(7,0.5740)(8,0.5758)(9,0.5779)(10,0.5794)(11,0.5782)(12,0.5770)(13,0.5761)(14,0.5758)(15,0.5756)(16,0.5746)(17,0.5749)(18,0.5745)(19,0.5740)(20,0.5740)(21,0.5733)(22,0.5739)(23,0.5754)(24,0.5776)(25,0.5786)(26,0.5796)(27,0.5805)(28,0.5821)(29,0.5818)(30,0.5844)};
    
    \addplot[name path=us_down,color=mycolor4!70] coordinates {(1,0.5244)(2,0.5380)(3,0.5513)(4,0.5529)(5,0.5556)(6,0.5511)(7,0.5481)(8,0.5439)(9,0.5414)(10,0.5423)(11,0.5430)(12,0.5451)(13,0.5464)(14,0.5461)(15,0.5454)(16,0.5442)(17,0.5440)(18,0.5429)(19,0.5411)(20,0.5408)(21,0.5392)(22,0.5394)(23,0.5404)(24,0.5411)(25,0.5412)(26,0.5412)(27,0.5405)(28,0.5404)(29,0.5398)(30,0.5404)};
    
    \addplot[mycolor4!50,fill opacity=0.5] fill between[of=us_top and us_down];

    \addplot[color=mycolor3] coordinates {(1,0.5324)(2,0.5344)(3,0.5378)(4,0.5353)(5,0.5385)(6,0.5429)(7,0.5387)(8,0.5425)(9,0.5409)(10,0.5339)(11,0.5272)(12,0.5245)(13,0.5266)(14,0.5237)(15,0.5211)(16,0.5191)(17,0.5185)(18,0.5245)(19,0.5271)(20,0.5278)(21,0.5245)(22,0.5260)(23,0.5292)(24,0.5288)(25,0.5264)(26,0.5272)(27,0.5288)(28,0.5341)(29,0.5390)(30,0.5401)};
    \addplot[name path=us_top,color=mycolor3!70] coordinates {(1,0.5346)(2,0.5455)(3,0.5514)(4,0.5448)(5,0.5488)(6,0.5555)(7,0.5526)(8,0.5515)(9,0.5477)(10,0.5400)(11,0.5327)(12,0.5298)(13,0.5334)(14,0.5316)(15,0.5306)(16,0.5286)(17,0.5282)(18,0.5367)(19,0.5415)(20,0.5419)(21,0.5378)(22,0.5396)(23,0.5448)(24,0.5468)(25,0.5466)(26,0.5489)(27,0.5531)(28,0.5578)(29,0.5612)(30,0.5633)};
    \addplot[name path=us_down,color=mycolor3!70] coordinates {(1,0.5301)(2,0.5233)(3,0.5242)(4,0.5259)(5,0.5282)(6,0.5303)(7,0.5248)(8,0.5336)(9,0.5340)(10,0.5279)(11,0.5218)(12,0.5193)(13,0.5199)(14,0.5158)(15,0.5115)(16,0.5095)(17,0.5088)(18,0.5123)(19,0.5127)(20,0.5136)(21,0.5113)(22,0.5124)(23,0.5136)(24,0.5107)(25,0.5062)(26,0.5056)(27,0.5046)(28,0.5105)(29,0.5168)(30,0.5168)};
    \addplot[mycolor3!50,fill opacity=0.5] fill between[of=us_top and us_down];
    \addlegendentry{(2,2)}
    \addlegendentry{(-2,-2)}
    
    \end{axis}
\end{tikzpicture} 
            \caption[Network2]%
            {{\small PROTEINS}}    
        \end{subfigure}
        \hspace{0.5cm}
        \begin{subfigure}{0.24\textwidth}
        \setlength 
        \figureheight{0.95in}
        \setlength 
        \figurewidth{1.4in}
        \centering  \scriptsize
        
        \definecolor{mycolor1}{rgb}{1,0.65,0}%
\definecolor{mycolor2}{rgb}{1,0.94,0}%
\definecolor{mycolor3}{rgb}{1.0, 0.25, 0.25}%
\definecolor{mycolor4}{rgb}{0.39, 0.58, 0.93}
\definecolor{mycolor5}{rgb}{0., 0.18, 0.39}

\begin{tikzpicture}
    \begin{axis}[
    width=0.951\figurewidth,
    height=\figureheight,
    at={(0\figurewidth,0\figureheight)},
    scale only axis,
    scaled x ticks=true,
    xticklabels={0,,20,,40,,60,,80,,100},
    xtick={0,4,8,12,16,20,24,28,32,36,40},
    x label style={at={(axis description cs:0.5,0.1)},anchor=north},
    xlabel style={font=\color{white!15!black},font=\fontsize{7.5}{7.5}\selectfont},
    xlabel={Iterations percentage(\%)},
    y label style={at={(axis description cs:0.22,0.5)},anchor=south},
    ylabel style={font=\color{white!15!black},,font=\fontsize{7.5}{7.5}\selectfont},
    ylabel={Sparsity},
      xmin=1, xmax=40,
      ymin=0.55, ymax=0.8,
      ymajorgrids=true,
      grid style=dashed,
      legend columns = 2,
      legend style={font=\fontsize{10}{10}\selectfont, legend cell align=left, align=left, draw=white!15!black, nodes={scale=0.7}, at={(0.85, 1.2)}},
      width=0.9\textwidth,
    ]

    \addplot[color=mycolor4] coordinates {(1,0.6661)(2,0.6631)(3,0.6610)(4,0.6650)(5,0.6619)(6,0.6604)(7,0.6618)(8,0.6618)(9,0.6617)(10,0.6625)(11,0.6637)(12,0.6659)(13,0.6693)(14,0.6717)(15,0.6735)(16,0.6757)(17,0.6761)(18,0.6763)(19,0.6877)(20,0.6979)(21,0.7011)(22,0.7063)(23,0.7138)(24,0.7187)(25,0.7277)(26,0.7283)(27,0.7284)(28,0.7295)(29,0.7273)(30,0.7242)(31,0.7197)(32,0.7172)(33,0.7163)(34,0.7137)(35,0.7138)(36,0.7160)(37,0.7166)(38,0.7156)(39,0.7177)(40,0.7188)};
    
    \addplot[color=mycolor3] coordinates {(1,0.6295)(2,0.6268)(3,0.6257)(4,0.6262)(5,0.6214)(6,0.6155)(7,0.6201)(8,0.6159)(9,0.6153)(10,0.6199)(11,0.6224)(12,0.6210)(13,0.6203)(14,0.6194)(15,0.6200)(16,0.6205)(17,0.6188)(18,0.6194)(19,0.6192)(20,0.6188)(21,0.6185)(22,0.6186)(23,0.6193)(24,0.6199)(25,0.6210)(26,0.6222)(27,0.6235)(28,0.6243)(29,0.6244)(30,0.6252)(31,0.6265)(32,0.6262)(33,0.6247)(34,0.6244)(35,0.6253)(36,0.6262)(37,0.6260)(38,0.6261)(39,0.6267)(40,0.6268)};

    \addplot[name path=us_top,color=mycolor4!70] coordinates {(1,0.7041)(2,0.7015)(3,0.7002)(4,0.7083)(5,0.7044)(6,0.7014)(7,0.7014)(8,0.7014)(9,0.7018)(10,0.7032)(11,0.7048)(12,0.7076)(13,0.7121)(14,0.7150)(15,0.7172)(16,0.7193)(17,0.7196)(18,0.7196)(19,0.7258)(20,0.7325)(21,0.7348)(22,0.7388)(23,0.7440)(24,0.7477)(25,0.7561)(26,0.7554)(27,0.7548)(28,0.7553)(29,0.7534)(30,0.7512)(31,0.7480)(32,0.7462)(33,0.7456)(34,0.7442)(35,0.7441)(36,0.7459)(37,0.7463)(38,0.7453)(39,0.7470)(40,0.7477)};
    
    \addplot[name path=us_down,color=mycolor4!70] coordinates {(1,0.6280)(2,0.6248)(3,0.6219)(4,0.6216)(5,0.6193)(6,0.6194)(7,0.6221)(8,0.6222)(9,0.6215)(10,0.6219)(11,0.6225)(12,0.6241)(13,0.6264)(14,0.6283)(15,0.6298)(16,0.6322)(17,0.6326)(18,0.6330)(19,0.6495)(20,0.6633)(21,0.6675)(22,0.6737)(23,0.6836)(24,0.6897)(25,0.6993)(26,0.7011)(27,0.7020)(28,0.7037)(29,0.7013)(30,0.6972)(31,0.6914)(32,0.6882)(33,0.6869)(34,0.6832)(35,0.6835)(36,0.6861)(37,0.6869)(38,0.6859)(39,0.6884)(40,0.6899)};
    
    \addplot[mycolor4!50,fill opacity=0.5] fill between[of=us_top and us_down];

    \addplot[name path=us_top,color=mycolor3!70] coordinates {(1,0.6741)(2,0.6700)(3,0.6692)(4,0.6733)(5,0.6677)(6,0.6621)(7,0.6642)(8,0.6647)(9,0.6610)(10,0.6651)(11,0.6675)(12,0.6648)(13,0.6641)(14,0.6632)(15,0.6640)(16,0.6649)(17,0.6624)(18,0.6631)(19,0.6624)(20,0.6618)(21,0.6616)(22,0.6618)(23,0.6629)(24,0.6637)(25,0.6654)(26,0.6672)(27,0.6689)(28,0.6701)(29,0.6701)(30,0.6706)(31,0.6719)(32,0.6712)(33,0.6694)(34,0.6690)(35,0.6703)(36,0.6713)(37,0.6711)(38,0.6714)(39,0.6719)(40,0.6720)};
    \addplot[name path=us_down,color=mycolor3!70] coordinates {(1,0.5849)(2,0.5837)(3,0.5821)(4,0.5791)(5,0.5751)(6,0.5689)(7,0.5760)(8,0.5671)(9,0.5695)(10,0.5747)(11,0.5773)(12,0.5771)(13,0.5766)(14,0.5756)(15,0.5759)(16,0.5761)(17,0.5752)(18,0.5758)(19,0.5760)(20,0.5759)(21,0.5755)(22,0.5753)(23,0.5757)(24,0.5761)(25,0.5765)(26,0.5772)(27,0.5781)(28,0.5785)(29,0.5787)(30,0.5797)(31,0.5810)(32,0.5813)(33,0.5801)(34,0.5797)(35,0.5803)(36,0.5810)(37,0.5808)(38,0.5808)(39,0.5814)(40,0.5817)};
    \addplot[mycolor3!50,fill opacity=0.5] fill between[of=us_top and us_down];
    
    \end{axis}
\end{tikzpicture} 
            \caption[Network2]%
            {{\small RDT-B}}    
        \end{subfigure}
        
        \caption{Evolution of explanation sparsity of three datasets during training.} 
        \label{sparsity}
\end{figure}

The explanation extracted from the model should become sparser as training goes on, since ENGAGE gradually removes superfluous information from representations. Thus, we define explanation \textit{sparsity}~\cite{pope2019explainability,yuan2020explainability} for graph $G_n$ as
$
    \mathcal{S}_n =1-(\sum_{i=1}^{\left|\mathcal{V}_n\right|}\mathds{1}{[\psi_{i} > \mu_{\psi}]})/\left|\mathcal{V}_n\right|,
$
where $\left|\mathcal{V}_n\right|$ is the number of nodes in $G_n$. $\psi_{i}$ is the importance score of node $v_i$. $\mu_{\psi}$ is the mean importance score of all nodes in graphs $\mathcal{G}$. The changes of explanation sparsity on DD, PROTEINS, and RDT-B at different training iterations are shown in Figure~\ref{sparsity}. We can observe that if we choose to perturb the original view radically ($\lambda_e=2$, $\lambda_f=2$), then sparsity tends to increase during the training process. In this mode, only the most crucial graph components will be kept, while other information will be discarded. If we perturb graphs softly ($\lambda_e=-2$, $\lambda_f=-2$), then the sparsity will remain stable or even decrease slightly, which means most of graph information is kept intact during training. We also provide visualization and quantitative analysis for explanation results of SAM in Appendix H and I, respectively. 

\begin{table}[t]
  \renewcommand\arraystretch{1.3}
  \caption{Effect of proposed SAM guided data augmentation.}
  \label{SAM_classificaiotn}
  \centering
\resizebox{0.99\columnwidth}{!}{
\begin{tabular}{l|cccc|cccc}
\hline
Method          & NCI1                  & PROTEINS              & DD     &PTC-MR          & COLLAB      & RDT-B       & RDT-M5K     & IMDB-B \\\hline\hline
RD-SimCLR           & 79.02 ± 0.52          &74.61 ± 0.56           &77.40 ± 0.81 &58.55 ± 2.01    &69.32 ± 1.54     &85.67 ± 5.40          &55.52 ± 0.74       &69.08 ± 2.47  \\
RD-Simsiam             &79.62 ± 0.59          &75.42 ± 0.50           &77.20 ± 1.33    &58.55 ± 	1.85              &66.50 ± 2.31 &86.12 ± 1.80          &54.30 ± 0.64       &69.86 ± 0.86  \\ \hline
HG-SimCLR           &80.48 ± 0.44  &74.93 ± 0.62  &78.40 ± 0.69 &56.62 ± 3.16  &72.53 ± 0.97  &89.92 ± 0.53        &56.01 ± 0.25 &69.88 ± 2.29  \\ 
HG-Simsiam          &80.70 ± 0.64  &75.45 ± 0.72  &77.87 ± 1.10 &58.61 ± 1.99  &72.84 ± 0.91 &89.63 ± 0.55 &55.92 ± 0.60 &71.28 ± 0.82  \\ \hline
EG-SimCLR       &\textbf{82.97 ± 0.20}           &75.44 ± 0.65     &\textbf{78.86 ± 0.51}     &61.51 ± 2.41                     &\textbf{76.60 ± 1.26}                   &\textbf{90.70 ± 0.46}          &56.22 ± 0.56       &71.78 ± 0.34  \\ 
EG-Simsiam       & 81.49 ± 0.19        &\textbf{76.06 ± 0.51}         &78.35 ± 0.83          &\textbf{62.67 ± 1.50}           &74.73 ± 0.79                  &89.17 ± 0.43          &\textbf{56.37 ± 0.17}      &\textbf{71.93 ± 0.40} \\ \hline
\end{tabular}}
\end{table}

\begin{figure}[t]
\label{heatmap}
\centering
\begin{subfigure}{0.3\textwidth}
     \includegraphics[width=\textwidth]{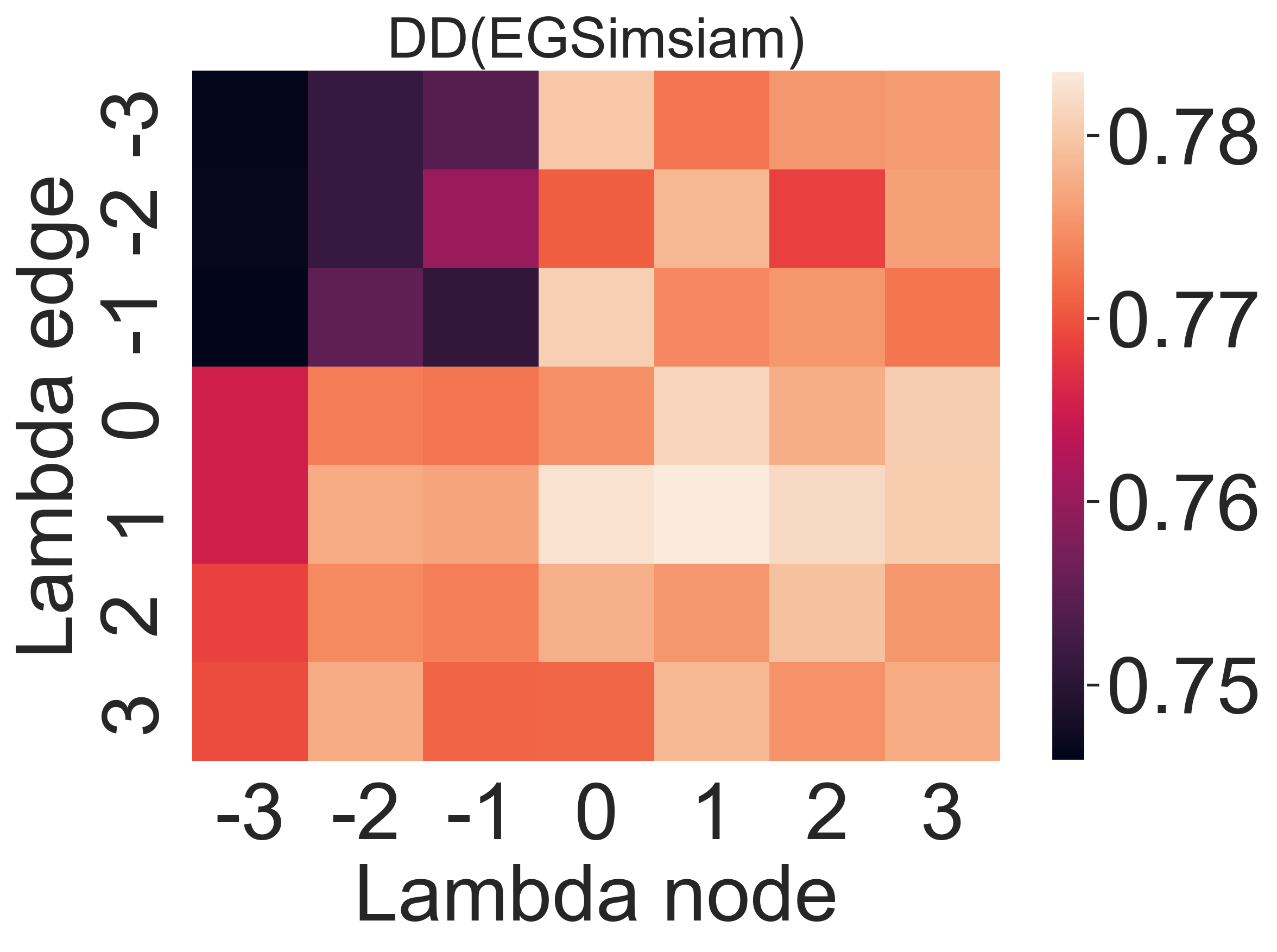}
    \caption{DD}
    \label{fig:first}
\end{subfigure}
\hspace{0.1cm}
\begin{subfigure}{0.3\textwidth}
    \includegraphics[width=\textwidth]{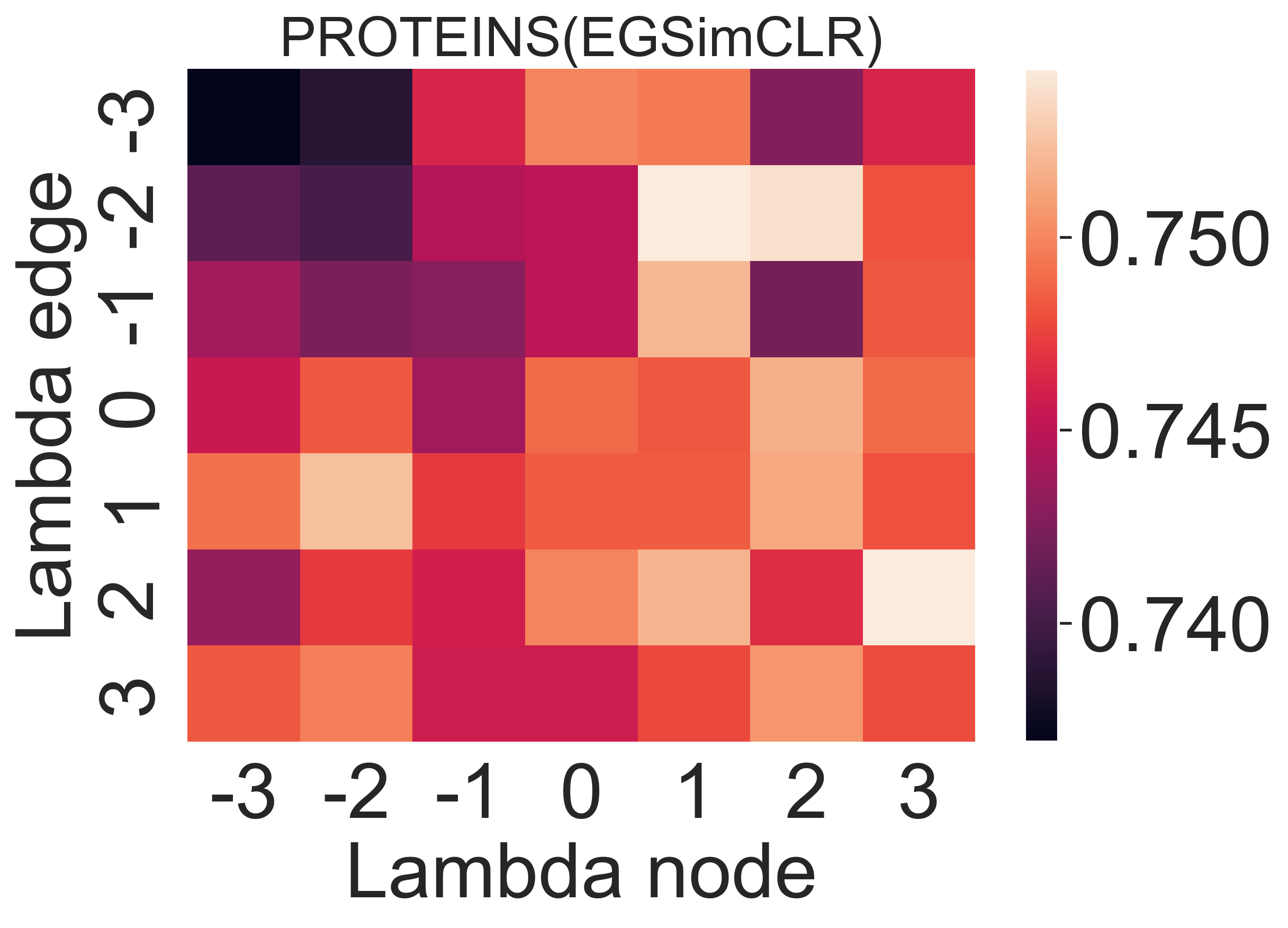}
    \caption{PROTEINS}
    \label{fig:second}
\end{subfigure}
\hspace{0.1cm}
\begin{subfigure}{0.3\textwidth}
    \includegraphics[width=\textwidth]{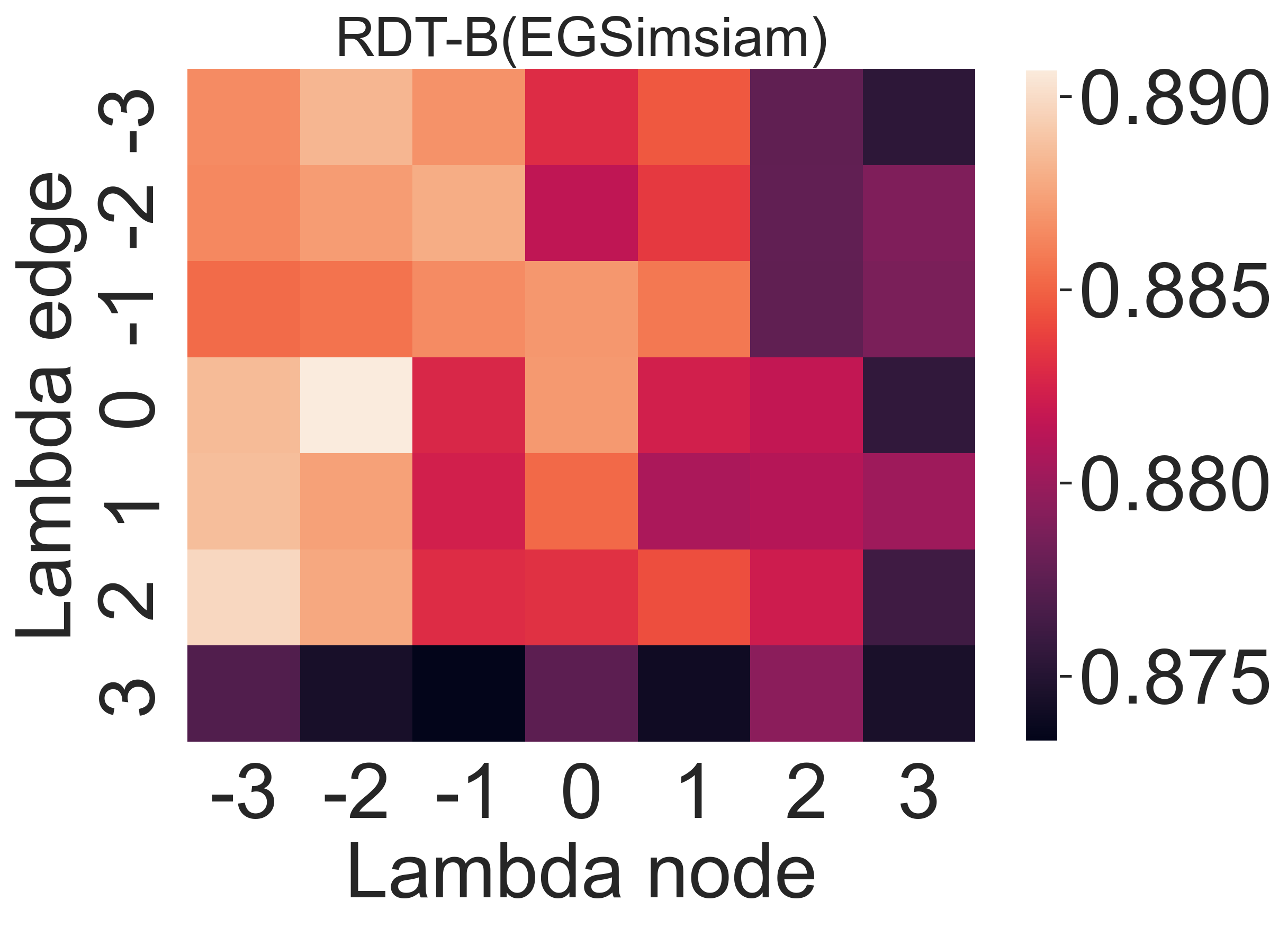}
    \caption{RDT-B}
    \label{fig:third}
\end{subfigure}

\caption{Model performances with different combinations of $\lambda_e$ and $\lambda_f$.}
\label{ablation}
\vspace{-10pt}
\end{figure}

\subsubsection{SAM guided data augmentation.} We first compare our proposed EG-SimCLR, EG-Simsiam models with RD-SimCLR and RD-Simsiam using random perturbation. Besides that, we formulate two variants of our method, called HG-SimCLR and HG-Simsiam ("HG" refers to "heatmap guided"), which do not employ local smoothing to obtain explanations.
The comparison is in Table~\ref{SAM_classificaiotn}. The result shows that the explanation guided models (HG and EG models) generally perform better than models with random data augmentation, which means the explanations can be successfully leveraged to improve representation learning with contrastive views. In addition, it can be observed that our proposed SAM-based methods consistently outperform other baselines, showing that local smoothing helps capture the important information more accurately. 

\subsubsection{Hyperparameter Analysis.} We now explore the effect of different $(\lambda_e, \lambda_f)$ combinations on model performance. We conduct a detailed study on 13 datasets, where $\lambda_e$ and $\lambda_f$ are changed, and all the other settings are the same. Larger $\lambda$ means a greater portion of data is perturbed. The results of DD, PROTEINS, and RDT-B datasets are shown in Figure~\ref{ablation}. The results on other datasets are provided in Appendix C.

Some observations are made as follows. 
\underline{First}, the optimal data augmentation threshold varies for different datasets. For example, Figure~\ref{ablation}(a) and (c) show that DD benefits from more perturbation, but RDT-B does not. The reason could be that, without data augmentation and contrastive learning, the vanilla representations in different datasets contain different levels of superfluous information.
\underline{Second}, the sensitivity to $\lambda_e$ and $\lambda_f$ varies across different datasets. Here we define performance gap as the difference between the best accuracy and the worst one. For graph-level and node-level tasks, COLLAB and CiteSeer have a maximum performance gap of 9.55\% and 2.45\%, respectively, while the RDT-M5K and Amazon-Photo have a minimum performance gap of 0.97\% and 0.23\%. The possible reason could be that in some datasets, the task-relevant information is redundant, so it does not hurt performance when more graph components are removed. We thus benefit more from searching for the optimal hyperparameters for datasets like COLLAB and CiteSeer.
\underline{Third}, Simsiam-based models generally show better stability than SimCLR-based models. The average performance gap of Simsiam-based models is 3.39\% and 0.75\% in graph-level and node-level tasks, respectively. By comparison, the corresponding numbers are 3.96\% and 1.08\% for SimCLR ones.

\section{Conclusion and Future Work}
In this paper, we propose an explanation guided data augmentation methods for graph representation learning. First, we design a new explanation method for interpreting unsupervised graph learning models by leveraging the information shared by local embeddings. Then we propose guided data augmentation using explanation results. We will preserve the graph structure and features whose importance exceeds the importance threshold. For the other part, we try to keep the mutual information between two views as low as possible. The final results show that our proposed methods achieve superior performance on multiple datasets. The ablation study also shows that the optimal thresholds vary across different datasets, while a general random perturbing may hurt task performance. For future work, we will (1) design more faithful unsupervised GNN explanation methods; (2) design intelligent explanation guided augmentation methods for other applications with domain knowledge.

\section{Acknowledgements}
The work is in part supported by NSF grant IIS-2223768. The views and conclusions contained in this paper are those of the authors and should not be interpreted as representing any funding agencies.

\newpage

\section*{Ethical Statement}

Our team acknowledges the importance of ethical considerations in the development and deployment of our ENGAGE framework. We ensure that our work does not lead to any potential negative societal impacts. We only use existing datasets and cite the creators of those datasets to ensure they receive proper credit. Additionally, we do not allow our work to be used for policing or military purposes. We believe it is essential to prioritize ethical considerations in all aspects of machine learning and data mining to ensure that these technologies are used for the benefit of society as a whole.
\newpage

\bibliographystyle{splncs04}
\bibliography{ecml}
\end{document}